\documentclass[11pt]{article}

\usepackage[T1]{fontenc}
\usepackage[utf8]{inputenc}

\usepackage[a4paper,margin=2.5cm]{geometry}

\usepackage{graphicx}
\usepackage{booktabs}
\usepackage{multirow}

\usepackage{enumitem}

\usepackage{hyperref}

\usepackage[accsupp]{axessibility}

\usepackage{authblk}

\usepackage{orcidlink}

\title{Contrastive-to-Self-Supervised: A Two-Stage Framework for Script Similarity Learning}

\author[1]{Claire Roman\,\orcidlink{0009-0001-5902-6606}}
\author[2]{Philippe Meyer\,\orcidlink{0000-0002-0618-2947}}

\affil[1]{{\normalsize University of Haute Alsace, IRIMAS UR 7499, Mulhouse, France}}
\affil[2]{{\normalsize Université Paris-Saclay, INRAE, AgroParisTech, Micalis Institute, Jouy-en-Josas, France}}

\date{}

\begin{document}

\maketitle

\vspace{-0.5cm}

\begin{abstract}
Learning similarity metrics for glyphs and writing systems faces a fundamental challenge: while individual graphemes within invented alphabets can be reliably labeled, the historical relationships between different scripts remain uncertain and contested. We propose a two-stage framework that addresses this epistemological constraint. First, we train an encoder with contrastive loss on labeled invented alphabets, establishing a teacher model with robust discriminative features. Second, we extend to historically attested scripts through teacher-student distillation, where the student learns unsupervised representations guided by the teacher's knowledge but free to discover latent cross-script similarities. The asymmetric setup enables the student to learn deformation-invariant embeddings while inheriting discriminative structure from clean examples. Our approach bridges supervised contrastive learning and unsupervised discovery, enabling both hard boundaries between distinct systems and soft similarities reflecting potential historical influences. Experiments on diverse writing systems demonstrate effective few-shot glyph recognition and meaningful script clustering without requiring ground-truth evolutionary relationships.
\end{abstract}

\noindent\textbf{Keywords:} 
Contrastive representation learning; teacher--student distillation; few-shot visual recognition.

\section{Introduction}
\label{sec:intro}

Ancient writing systems encode millennia of human history, yet their visual relationships are still not fully understood. Deciphering whether two scripts share a common ancestor, or whether a glyph form migrated across cultures, is a question that archaeologists and linguists have debated for decades \cite{daniels1996world}. Computational approaches offer a promising route: if a model can learn a geometrically coherent similarity space over glyphs and scripts, it could provide objective, reproducible evidence to inform these debates. Yet this goal faces a fundamental conceptual challenge, one that, to our knowledge, no existing visual representation learning framework has explicitly addressed.

The obstacle is asymmetric supervision. At the glyph level, learning from multiple instances of the same character poses no conceptual difficulty: different renderings of the same glyph can safely be treated as positives, allowing models to learn deformation-invariant representations. Likewise, when working with invented alphabets whose characters are historically independent by design, treating different glyph classes as negatives is unproblematic. Indeed, many metric-learning approaches implicitly assume such a setting and construct negative pairs between all distinct character classes \cite{koch2015siamese}. Historical writing systems, however, do not satisfy these assumptions. Characters within a script may share structural motifs, and visual similarities may reflect historical influence, borrowing, or common graphic conventions. More importantly, relationships between scripts are often uncertain, debated, or only partially documented~\cite{parpola86indus, duhan2022origin}. Defining negative pairs across scripts by asserting that two characters are unrelated therefore risks baking in unverifiable linguistic assumptions.

This paper addresses the asymmetric supervision problem by separating what can be supervised from what must remain exploratory. We propose a two-stage framework that first learns a strong discriminative prior from labeled invented alphabets, scripts whose character identities are unambiguous and historically independent, and then adapts this prior to historically attested scripts without imposing cross-script negatives. In Stage~1, a teacher encoder is trained with supervised contrastive learning~\cite{khosla2020supervised} on fictional scripts producing an embedding space with clean intra-class clustering and inter-class separation. In Stage~2, this structure is transferred to unlabeled historical scripts through an asymmetric teacher-student distillation objective inspired by BYOL~\cite{byol}: the student is encouraged to match a momentum-updated target network under genuine handwriting variability, allowing representations to reorganize where needed while preserving the teacher's geometric regularities.

Our evaluation reflects the dual objective of the task. At the glyph level, we assess few-shot recognition via 20-way 1-shot retrieval. At the script level, we induce script-to-script distances by aggregating nearest-neighbor glyph matches and evaluate the resulting rankings against curated linguistic similarity levels using NDCG@10 and Spearman correlation. Across five backbone architectures, our hybrid training consistently achieves the best NDCG@10, our primary metric for script-level ranking quality, over all purely self-supervised baselines, while remaining competitive on glyph-level retrieval. Qualitatively, this shows that the student not only inherits the teacher's discriminative structure, but accentuates historically grounded proximities.

In summary, our contributions are: \textbf{(i)} a two-stage training strategy that decouples reliable character supervision from uncertain script relations; \textbf{(ii)} a teacher-initialized self-distillation adaptation that avoids cross-script negatives while allowing representations to reorganize on historical data; and \textbf{(iii)} an evaluation protocol combining few-shot glyph retrieval and script-level ranking metrics, validated on Omniglot and a newly constructed Unicode dataset.

\section{Related Work}

\subsubsection{Few-Shot and Metric Learning.}

The Omniglot dataset has played a central role in character-level modeling, serving both as a benchmark and as a training corpus for visual representation learning on handwritten characters \cite{lake2015human,lake2019omniglot}. Building on this setting, metric-based approaches such as siamese networks \cite{koch2015siamese, liu2022one}, matching networks \cite{vinyals2016matching}, and prototypical networks \cite{snell2017prototypical} have been trained on labeled character data for one-shot recognition. These learned glyph embeddings have subsequently enabled quantitative analyses at the script level \cite{roman2024analysis}.

\subsubsection{Contrastive and Self-Supervised Representation Learning.}

Contrastive learning methods such as SimCLR \cite{chen2020simple} and SupCon \cite{khosla2020supervised} learn representations by attracting positive pairs while explicitly repelling all remaining samples, treated as negatives by default. This mechanism produces well-clustered embedding spaces but introduces an inherent tension: samples that are not designated as positives are uniformly pushed away, regardless of their actual semantic proximity \cite{wang2020understanding}. Self-supervised methods such as Barlow Twins \cite{barlow} address this limitation by abandoning the notion of negatives entirely: rather than defining a repulsion term, they encourage the cross-correlation matrix between two augmented views of the same sample to be as close to the identity as possible, shaping the embedding space through redundancy reduction alone. Beyond benchmark settings, unsupervised deep learning has also been applied to ancient corpora with debated sign inventories \cite{corazza2022unsupervised}.


\subsubsection{Knowledge Distillation and Self-Distillation.}

Many recent representation learning methods rely on teacher-student training to stabilize optimization and transfer structure. Classical knowledge distillation \cite{hinton2015distilling} trains a student to match a teacher's softened outputs, enabling compact models to inherit information learned under stronger supervision. In self-distillation, the teacher is typically a momentum-updated version of the student. BYOL \cite{byol} exemplifies this paradigm by learning invariant features through prediction of an EMA teacher without requiring negative pairs. Related approaches such as DINO \cite{caron2021dino} and its large-scale extension DINOv2 \cite{oquab2023dinov2} further show that strong visual representations can emerge from teacher-student training without manual annotations.

\section{Methodology}

\subsection{Motivation for the Two-Stage Approach}\label{method}

The fundamental challenge in studying similarities between ancient graphemes and writing systems lies in the absence of reliable ground truth regarding historical influences and evolutionary relationships. While we can confidently label individual glyphs within invented alphabets, such as the alien alphabet appearing in Matt Groening's animated series Futurama or the Tengwar script developed by J.R.R. Tolkien for the elvish languages Quenya and Telerin, we cannot make definitive claims about whether two distinct characters of historical writing systems should be considered similar or dissimilar without risking the introduction of unverifiable archaeological or linguistic assumptions. This epistemological constraint motivates our two-stage learning framework.

\subsubsection{Stage 1: Supervised Contrastive Learning on Labeled Invented Glyphs.}

We first train a deep encoder (including standard CNN, Siamese, and Resnet) with contrastive loss on a curated set of invented glyphs whose ground truth labels do not introduce bias toward historical writing systems. Here, each character and its variations is considered as its own class. In addition to these handwritten variations, data augmentation (rotation, shear, zoom, translation) has been performed to preserve class identity \cite{koch2015siamese}. Different characters from these alphabets belong to distinct classes. This supervised phase produces a teacher model that learns robust feature representations capable of distinguishing between writing systems with known independence. Following \cite{khosla2020supervised}, we adopt the "many-to-many" supervised contrastive loss (referred to as $\mathcal{L}^{out}$ in the original paper), which considers all instances of the same class as positives, making it particularly suited to our setting where each character class contains multiple handwritten variations and augmented samples. Glyphs from distinct classes act implicitly as negative pairs. Given a batch of $B$ glyph images with $\ell_2$-normalized embeddings $\{\mathbf{z}_i\}_{i=1}^B$ and labels $\{y_i\}_{i=1}^B$ where $y_i \in \{0,\dots,G-1\}$ is a discrete label indicating the glyph identity of sample $i$, and $G = 350$ denotes the number of distinct glyph classes in the dataset of invented alphabets, the supervised contrastive loss is defined as
\begin{equation}
\mathcal{L}_{\mathrm{sup}}
=
\frac{1}{|\mathcal{I}|}
\sum_{i \in \mathcal{I}}
\ell_i,
\end{equation}
where $\mathcal{I} = \{ i \mid \exists\, j \neq i \text{ such that } y_j = y_i \}$
denotes the set of anchors with at least one positive, and
\begin{equation}
\ell_i
=
-
\frac{1}{|\mathcal{P}(i)|}
\sum_{p \in \mathcal{P}(i)}
\log
\frac{
\exp\!\left(\mathbf{z}_i^\top \mathbf{z}_p / \tau\right)
}{
\sum\limits_{\substack{a\in A(i)}}
\exp\!\left(\mathbf{z}_i^\top \mathbf{z}_a / \tau\right)
}.
\end{equation}

\noindent where $A(i) = \mathcal{I} ~\backslash ~ \{i\}$, $\mathcal{P}(i) = \{ p \in A(i), ~ y_i = y_p \}$ and $\tau \in \mathbb{R}^*$ is a scalar temperature hyperparameter, which controls the concentration of the distribution in the embedding space: a low value of $\tau$ sharpens the distribution, penalising hard negatives more severely and encouraging tighter clusters, while a higher value produces a smoother, more uniform distribution over negatives. Upon convergence, the teacher encoder $f_{\phi}^*$ yields a geometrically structured embedding space where characters from distinct invented alphabets form well-separated clusters (see Figure \ref{fig:t_sne}). This discriminative structure serves as a semantic prior for the unsupervised adaptation stage described below. The architecture and hyperparameters of $f_{\phi}^*$ are detailed in Section \ref{experimental_results}. 
\subsubsection{Stage 2: Unsupervised Teacher-Student Distillation.}

However, the majority of historical writing systems cannot be cleanly separated into distinct classes due to unknown evolutionary relationships (e.g., potential influence of Phoenician on Greek) and incomplete historical records regarding script transmission. In this setting, we can only assert that augmented versions of the same character belong together, but we cannot reliably define negative pairs across historically attested scripts without introducing unverifiable linguistic assumptions. This motivates the use of a self-supervised framework that operates without explicit negative pairs \cite{byol}. 

\subsubsection{BYOL Framework.} Given an unlabeled dataset of glyph images $\mathcal{G}_u = \{x_i\}_{i=1}^{M}$ ($M = 922$), BYOL \cite{byol} relies on two networks sharing the same backbone architecture: an online network (student) $f_\theta$ paired with a predictor MLP $q_{\theta}$, and a target network $f_\xi$ updated via Exponential Moving Average (EMA) of the student weights: 
\begin{equation}
\xi = \kappa \cdot \xi + (1-\kappa) \cdot \theta, ~~ \kappa \in [0,1].
\end{equation}
For each image $x_i,~ i\in \{1,\ldots,M\}$ in a batch of size $B'$, two views $(x_i^1, x_i^2)$ are generated via data augmentation (geometric and photometric transformations). The student produces predictions through the predictor: $p_i^1 = q_{\theta}(f_{\theta}(x_i^1))$ and $p_i^2 = q_{\theta}(f_{\theta}(x_i^2))$ while the target network produces representations with a stop-gradient (sg): $z_i^1 = \text{sg}(f_{\xi}(x_i^1))$ and  $z_i^2 = \text{sg}(f_{\xi}(x_i^2))$. The training objective minimizes the symmetrized negative cosine similarity between student predictions and target representations:
\begin{equation}
\mathcal{L}_{\mathrm{BYOL}} = \frac{1}{B'} \sum_{i=1}^{B'}\left[\mathcal{D}(p_i^1, z_i^2) + \mathcal{D}(p_i^2, z_i^1)\right]
\end{equation}

\noindent where $\mathcal{D}(p, z) = 2 - 2 \cdot \frac{p^\top z}{\|p\|_2 \cdot \|z\|_2}$. The asymmetry between the student (with predictor) and the target network (without predictor, no gradient flow) prevents representational collapse in the embedding space without requiring negative pairs. 

\subsubsection{Our adaptation.} 
\begin{figure}[h!]
  \centering
  \includegraphics[width=0.95\columnwidth]{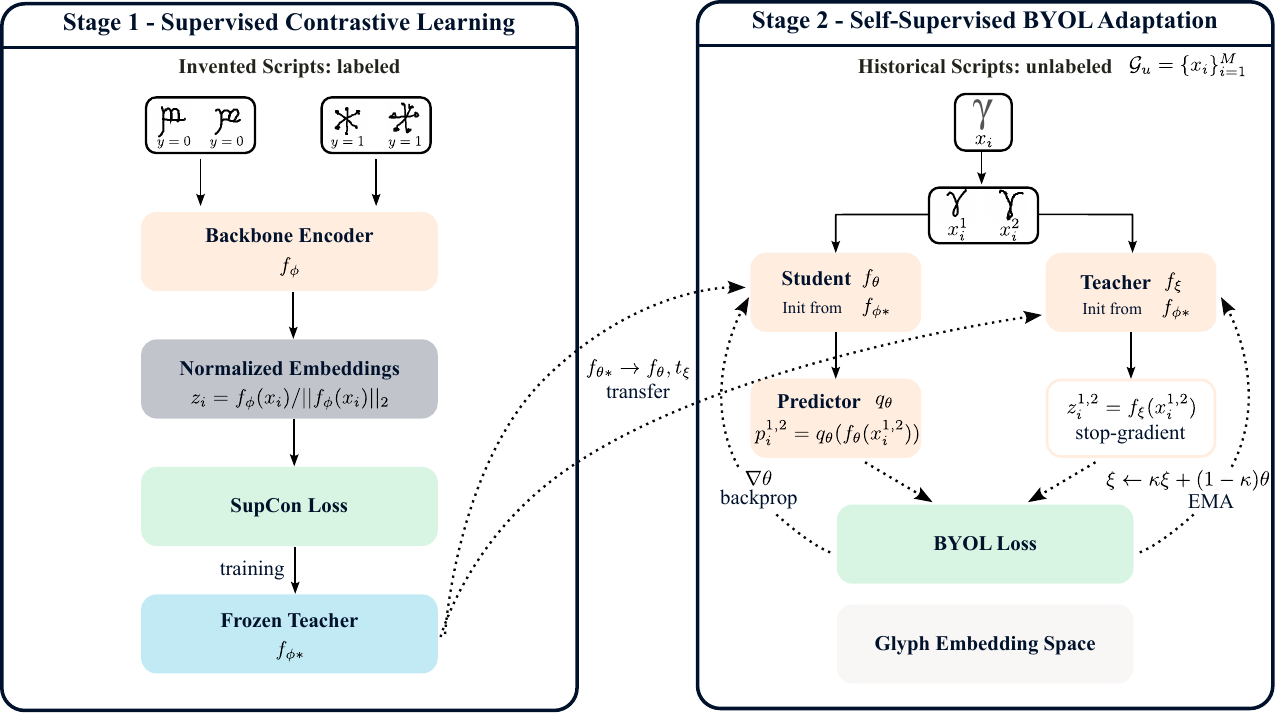}
\caption{Two-stage framework. Stage 1: train a teacher encoder $f_{\phi}^*$ with supervised contrastive learning (SupCon) on labeled invented scripts, yielding a discriminative embedding space. Stage 2: adapt to unlabeled historical scripts via teacher-initialized BYOL: a student $f_{\theta}$ with predictor $q_{\theta}$ matches a momentum (EMA) target network $f_{\xi}$ on two augmented views, using stop-gradient, and without cross-script negatives.}  \label{fig:architecture}
\end{figure}

Our approach departs from standard BYOL in three key ways, illustrated in Figure \ref{fig:architecture}.
\noindent First, rather than initializing both networks randomly, the student $f_{\theta}$ and the target network $f_{\xi}$ are both initialized from the teacher $f_{\phi}^*$ obtained at the end of Stage 1. This provides a semantically structured initialization that guides the student toward meaningful representations from the very first iterations, as opposed to standard BYOL which must discover representational structure from scratch without any semantic prior.

Second, regarding the network architecture, we deliberately omit the projection MLP typically inserted between the backbone and the predictor in the original BYOL formulation. Since our backbone already produces compact low-dimensional embeddings ($d \in \{64, 128, 256\}$, see \ref{exp_setup}), an additional projector would add unnecessary complexity without meaningful benefit, and would risk overfitting given the moderate size of the Omniglot dataset. The predictor $q_\theta$ thus operates directly on the backbone embeddings.

Third, rather than generating two augmented views from a single image 
as in standard BYOL, we leverage the availability of multiple genuine handwritten 
instances per character class, complemented by the same geometric augmentations 
(rotation, shear, zoom, translation) applied in Stage~1.

\subsection{Glyph and Script Similarity}

\subsubsection{Glyph-Level Similarity.}

Once the model has been trained, it defines an embedding space where each glyph $x_i$ is mapped to a $d$-dimensional feature vector $z_i$ and $\ell_2$-normalized. For any pair of glyph images $(x_1, x_2)$, we compute their similarity as the cosine of their embeddings, and their dissimilarity as follows:
\begin{equation} \label{eq:sim_glyph}
\mathrm{sim}(x_1, x_2) = \cos\big(z_1, z_2\big) = z_1^\top z_2, \qquad d_g(x_1,x_2) := 1 - \mathrm{sim}(x_1,x_2).
\end{equation}
This quantity reflects the angular proximity between glyph representations in the learned feature space. This procedure is applied independently for each trained model, thereby inducing a model-specific geometry on the space of glyphs.

\subsubsection{Script-Level Similarity.}

A script $s$ is represented as a finite set of glyphs. To compare two scripts $s_1$ and $s_2$, we aggregate glyph-level dissimilarities through a nearest-neighbor matching scheme. First, we define the directed average distance
\begin{equation}
\tilde{d}_s(s_1, s_2)
:=
\frac{1}{|s_1|}
\sum_{x_1 \in s_1}
\min_{x_2 \in s_2}
d_g(x_1, x_2),
\end{equation}
which measures how well each glyph of $s_1$ can be approximated by its closest counterpart in $s_2$. To obtain a symmetric script-level dissimilarity, we define
\begin{equation}\label{eq:dist_script}
d_s(s_1, s_2)
:=
\frac{1}{2}
\Big(
\tilde{d}_s(s_1, s_2)
+
\tilde{d}_s(s_2, s_1)
\Big).
\end{equation}
This formulation allows multiple glyphs from one script to be associated with the same glyph in another to reflect the historical evolution of writing systems \cite{roman2024analysis}.

\subsection{Evaluation Metrics}

\subsubsection{20-Way 1-Shot Evaluation.}

At the glyph level, representation quality is evaluated using a 20-way 1-shot classification task formulated as a retrieval problem \cite{lake2015human, koch2015siamese}. For each episode, $N=20$ character classes are randomly sampled, and one class is selected as the target. Two distinct images from this class are drawn: one serves as the query glyph $x_q$, and the other as the positive candidate. From each of the remaining $N-1$ classes, a single image is sampled to serve as a negative candidate. The similarity between the query and a candidate glyph $x$ is computed using Equation~\eqref{eq:sim_glyph} and candidates are ranked in decreasing order of similarity. The prediction is considered correct if the true positive candidate appears among the top-$k$ ranked glyphs. Performance is reported using Top-$k$ accuracy (with $k \in \{1,5\}$), averaged over 400 independently sampled episodes.

\subsubsection{Normalized Discounted Cumulative Gain.}

Beyond character-level discrimination, we evaluate whether the learned representations capture higher-level relationships between writing systems. Given a query script $s_q$, scripts are ranked by increasing $d_s(s_q,\cdot)$. Ground-truth inter-script relationships are encoded as discrete similarity levels derived from linguistic and historical classifications, where lower levels indicate stronger similarity. These levels are converted into graded relevance scores. The Discounted Cumulative Gain at rank $k$ and its normalized form are defined as
\begin{equation}
\mathrm{DCG@}k
=
\sum_{r=1}^{k}
\frac{\mathrm{rel}_r}{\log_2(r+1)}, 
\qquad
\mathrm{NDCG@}k
=
\frac{\mathrm{DCG@}k}{\mathrm{IDCG@}k},
\end{equation}
where $\mathrm{rel}_r$ denotes the relevance of the script ranked at position $r$. The logarithmic discount penalizes relevant scripts appearing lower in the ranking. Normalization by the Ideal DCG (IDCG), obtained from the ground-truth ordering, ensures comparability across queries. We report NDCG@10 averaged over all query scripts.

\subsubsection{Spearman’s Rank Correlation Coefficient.}

To quantify the global agreement between embedding-induced script geometry and ground-truth similarity levels, we compute Spearman’s rank correlation coefficient \cite{daniel1990applied}. For each pair of scripts $(s_i, s_j)$ in the evaluation set, we consider the predicted script dissimilarity $d_s(s_i, s_j)$ and the ground-truth similarity level $\ell(s_i, s_j)$. The Spearman’s coefficient $\rho$ measures the correlation between the ranks of predicted dissimilarities and ground-truth levels:
\begin{equation}
\rho
=
1
-
\frac{6 \sum_i (R_i - S_i)^2}{n(n^2 - 1)},
\end{equation}
where $R_i$ and $S_i$ denote the ranks of $d_s(s_i, s_j)$ and $\ell(s_i, s_j)$, respectively, across the $n$ evaluated script pairs. Since smaller similarity levels correspond to stronger relationships, a positive and statistically significant $\rho$ indicates that script dissimilarities $d_s$ increase consistently as linguistic similarity decreases. In other words, historically closer scripts tend to be embedded closer together.

\subsection{Datasets} \label{dataset}

\subsubsection{Omniglot.} \label{dataset_omniglot}

The Omniglot dataset \cite{lake2015human} consists of 1,623 handwritten character classes drawn from 50 different writing systems from around the world (details are given in the supplementary material). Each character class consists of 20 instances produced by different human subjects and each glyph is represented as a black-and-white $105 \times 105$ pixels image. To augment the dataset we generated $8$ additional instances per glyph instance via the random affine perturbations \cite{koch2015siamese}, each applied independently with probability $0.5$: rotation with angle $\theta \in [-10^\circ,10^\circ]$, shear with parameter $\rho \in [-0.3,0.3]$, zoom with factor $s \in [0.8,1.2]$, translation with pixel shifts $t_x,t_y \in [-2,2]$. The dataset contains historically attested scripts and fictional or modern alphabets introduced after the 18th century. We separate it in three distinct datasets (see Figure \ref{fig:data}): 
\begin{itemize}
    \item the Omniglot supervised invented dataset composed of the 15 fictional or modern scripts, dedicated for supervised learning,
    \item the Omniglot unsupervised dataset composed of 25 historically attested scripts, dedicated for unsupervised learning,
    \item the Omniglot evaluation dataset composed of 10 historically attested scripts.
\end{itemize}

\subsubsection{Unicode.}

\begin{figure}[h!]
  \centering
  \includegraphics[width=0.7\columnwidth]{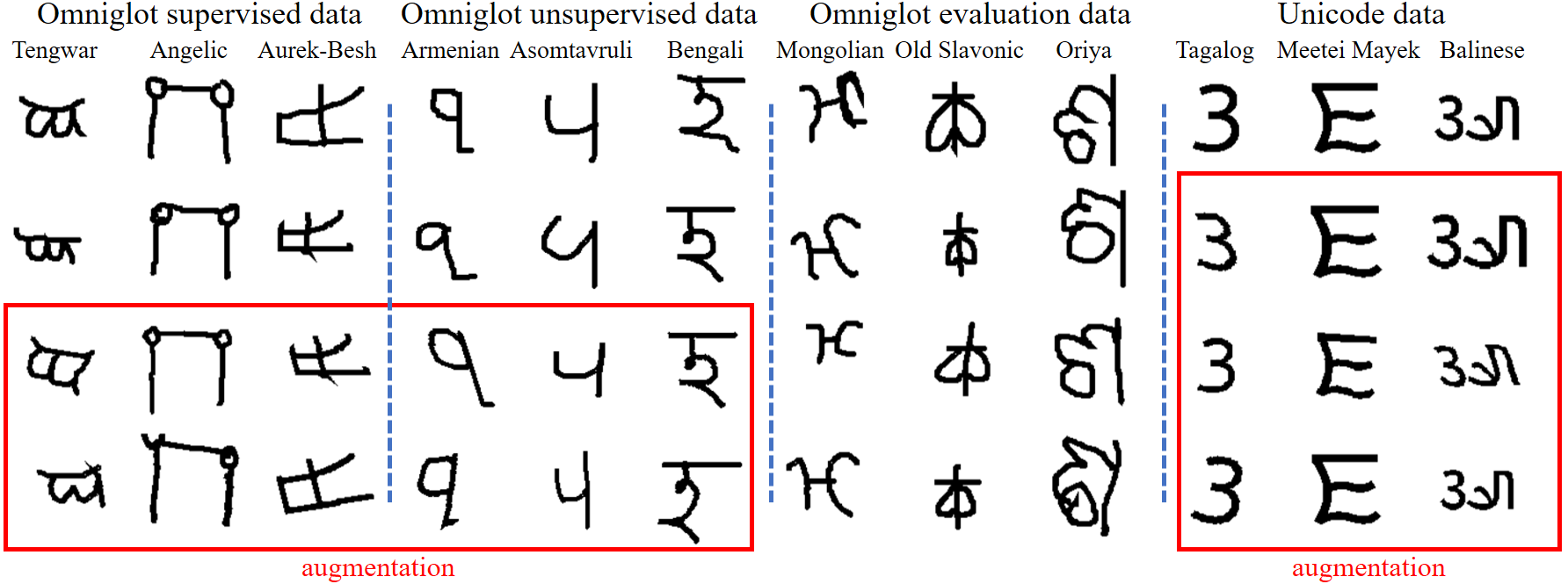}
\caption{Datasets used in this work. Omniglot is split into supervised (invented alphabets for Stage~1), unsupervised (historical scripts for Stage~2), and evaluation sets. A complementary Unicode character dataset is generated using Noto fonts. Augmented instances per glyph are obtained through random transformations (red boxes).}
\label{fig:data}
\end{figure}

To further evaluate the models we construct a glyph and script dataset by considering all writing systems whose historical development predates the 19th century from the Unicode 17.0 repertoire \cite{unicode17}. For each script, its character inventory is defined using explicit Unicode codepoint ranges and graphemes are rendered using the Google Noto font family \cite{google_noto}. This method offers broad and consistent coverage across scripts while limiting stylistic variation. Each Unicode character is centered using its non-white bounding box, with font size adaptively adjusted to prevent clipping and preserve comparable stroke widths, and rendered as a black-and-white $105 \times 105$ pixels image, see Figure \ref{fig:data}.

Ground-truth relationships between scripts are encoded as three discrete similarity levels derived from historical and typological knowledge of writing system evolution \cite{daniels1996world}. Level~1 (very similar) corresponds to scripts that are essentially variants of the same writing system or closely related historical forms, for example Georgian Asomtavruli and Georgian Mkhedruli, or Hiragana and Katakana. Level~2 (similar) groups scripts belonging to the same historical lineage or direct derivational branch, such as Greek and Cyrillic, or Brahmi and Devanagari. Level~3 (very different) corresponds to scripts from unrelated families with no clear historical derivation, for instance Latin and Hangul, or Devanagari and Chinese characters. Pairs not explicitly assigned to Levels~1--3 are treated as unrelated scripts and receive the lowest similarity level during evaluation to avoid introducing speculative historical relationships and permit to evaluate on the full set of script pairs. The complete list of Unicode scripts and their assigned similarity levels is provided in the supplementary material.

\section{Experimental Results} \label{experimental_results}

\subsection{Experimental Setup} \label{exp_setup}

\subsubsection{Baseline SotA.}

We evaluate our hybrid framework against several baselines designed to assess the contribution of each component of our approach. We compare against two competitive self-supervised methods trained without any labeled data. BYOL \cite{byol}, defined in section \ref{method} is a student-teacher framework that learns representations by predicting one augmented view from another using an EMA-updated target network, without requiring negative pairs. Barlow Twins \cite{barlow} learns representations by minimizing redundancy between the features of two augmented views via cross-correlation matrix regularization. Both methods serve as reference points to assess whether adding a supervised pre-training phase brings any benefit over purely self-supervised approaches. We further include DINOv2-ViT-S/14 \cite{oquab2023dinov2}, a 
self-supervised visual foundation model trained on 142M curated images, 
designed to produce all-purpose visual features that generalise across 
tasks and image distributions without fine-tuning. We evaluate it in two 
configurations: frozen features and fine-tuned projection head via BYOL 
on the unlabeled data. This baseline allows us to test whether a 
general-purpose foundation model is sufficient for ancient script analysis, 
or whether domain-specific training is necessary.

To isolate the contribution of Stage 2, we additionally evaluate the frozen teacher $f_\phi^*$ obtained at the end of Stage 1, without any subsequent BYOL adaptation. This ablation directly answers the question of whether the supervised contrastive pre-training alone is sufficient, or whether the unsupervised adaptation on historical scripts is necessary to obtain strong representations on the evaluation set. Additionaly, in our framework, Stage 2 produces two networks: the student $f_\theta$, trained by gradient descent, and the target network $f_\xi$, updated by EMA of the student weights. Since the EMA target network is known to produce smoother and more stable representations \cite{byol}, we report results for both networks at inference time, allowing us to assess the respective quality of each.

\subsubsection{Data Splits.} Following the dataset description in Section~\ref{dataset_omniglot}, Stage 1 uses the Omniglot supervised dataset (15 fictional and modern scripts, $G = 350$ character classes) with labeled instances. For the hybrid approach, Stage 2 uses the Omniglot unsupervised dataset (25 historically attested scripts) as unlabeled data. For the baselines (BYOL simple, Barlow Twins), both the supervised and unsupervised datasets are combined into a single unlabeled pool, ensuring a fair comparison, our hybrid approach has access to the same data overall. All methods are evaluated on the Omniglot evaluation dataset (10 historically attested scripts, unseen during training). For Stage 1, 10\% of characters are held out for validation at the character level to avoid class overlap.

\subsubsection{Evaluation Protocol.} We evaluate all methods using $N$-way 1-shot retrieval. For each episode, a query image is matched against $N = 20$ candidate images using cosine similarity in the $\ell_2$-normalized embedding space. We report Top-1 and Top-5 accuracy averaged over 400 episodes, as well as NDCG@10 and Spearman correlation to assess the quality of the full similarity ranking.

\subsubsection{Backbones.} We evaluate four backbone architectures of increasing capacity: a Simple CNN (2.3M parameters), a Siamese-like CNN (8.7M) (a branch of \cite{koch2015siamese}), ResNet-18 (11.7M), ResNet-34 (21.8M), and ResNet-50 (25.6M) \cite{he2016deep}. All backbones produce $\ell_2$-normalized embeddings of dimension $d \in \{64, 128, 256\}$, except the Siamese CNN which outputs $d = 4096$-dimensional embeddings following \cite{koch2015siamese}. We additionally evaluate DINOv2-ViT-S/14 \cite{oquab2023dinov2} (22M parameters) as a large pretrained baseline, both frozen and fine-tuned. Note that BYOL \cite{byol} and Barlow Twins \cite{barlow} were originally proposed with a ResNet-50 backbone; we therefore include ResNet-50 as a common reference point for these baselines alongside our hybrid approach. More details about the architectures are provided in the supplementary material.

\subsubsection{Hyperparameter Optimization.} All methods are tuned using Optuna \cite{optuna} with the TPE sampler and Median pruner over 30 trials, maximizing 1-NN accuracy on the validation set. The common search space covers batch size $\in \{128, 256, 512\}$, learning rate $\in [3 \times 10^{-5}, 3 \times 10^{-4}]$, weight decay $\in [10^{-7}, 10^{-4}]$, warmup epochs $\in [5, 60]$, and gradient clipping $\in [0, 2]$. Method-specific parameters include temperature $\tau \in [0.05, 0.3]$ and teacher epochs $\in [50, 300]$ for Hybrid (Stage~1), EMA decay $\in [0.95, 0.9995]$ and predictor hidden dimension $\in \{256, 512, 1024\}$ for BYOL-based methods, and $\lambda_{\text{off-diag}} \in [10^{-4}, 10^{-1}]$ with projection dimension $\in \{512, 1024, 2048, 4096\}$ for Barlow Twins.

Across models, the optimal configurations consistently favored:
\begin{itemize}
\item moderate batch sizes (128--256 in most cases, 512 for some ResNet variants),
\item learning rates mainly in the $10^{-4}$ range (sometimes lower for ResNet-based backbones; higher for some Stage~1 teachers),
\item very small weight decay values (typically $10^{-6}$--$10^{-7}$),
\item long training schedules (roughly 300--500 epochs when applicable),
\item high EMA decay values (0.95--0.999) for momentum-based methods,
\item projection/predictor hidden dimensions between 512 and 2048 with LR multipliers in the 1--4 range,
\item moderate gradient clipping (generally below 2.0),
\item method-specific parameters such as temperatures in the 0.05--0.22 range (Hybrid) and $\lambda_{\text{off-diag}}$ spanning $\approx 4 \times 10^{-4}$ to $4 \times 10^{-2}$ (Barlow Twins).
\end{itemize}

The complete list of best hyperparameters for every study is reported in the supplementary material.

\subsubsection{Implementation Details.} All models are trained with the AdamW optimizer using a warmup and cosine decay learning rate schedule. At each training step, a single pair $(x_1, x_2)$ of distinct handwritten instances is sampled per character class, leveraging the multiple genuine handwritten variations available in Omniglot rather than relying solely on synthetic augmentations.\\

\subsection{Benchmark}

\subsubsection{Primary Metrics: Script-Level Ranking Quality.}

The primary objective of our framework is not merely to discriminate between individual glyphs, but to recover a geometrically coherent embedding space that reflects the historical and linguistic relationships between writing systems. We therefore consider NDCG@10 and Spearman's rank correlation coefficient as our primary evaluation metrics, as they directly measure the quality of the script-level ranking induced by the learned representations. NDCG@10 quantifies whether historically related scripts are ranked among the closest neighbours of a query script, with a logarithmic discount that penalises relevant scripts appearing lower in the ranking. Spearman's coefficient $\rho$ measures the global monotonic agreement between embedding-induced dissimilarities and ground-truth linguistic similarity levels: a positive and statistically significant $\rho$ indicates that historically closer scripts are consistently embedded closer together in the learned feature space.

As shown in Table~\ref{table_results}, our hybrid approach achieves the highest NDCG@10 on three backbone architectures (Simple CNN, ResNet-34, and ResNet-50). The remaining cases are won by SupCon on the Siamese CNN and Barlow Twins on ResNet-18. This consistent superiority confirms that the semantic prior established during Stage 1, through supervised contrastive learning on invented alphabets, enables the model to organise the embedding space in a way that is more aligned with historical script relationships than purely self-supervised methods. Notably, our approach achieves the best NDCG@10 on ResNet-50 (0.3178), outperforming both Barlow Twins (0.2997) and BYOL (0.2708) by a significant margin on this backbone.

Regarding Spearman's correlation, our method achieves the best score on Simple CNN ($\rho = 0.640$) and competitive results on Siamese CNN ($\rho = 0.594$). These results confirm that the global ordering of script dissimilarities is well preserved in our embedding space. However, BYOL achieves higher Spearman scores on ResNet-based architectures, suggesting that on larger backbones, the EMA-based self-supervised signal alone may produce smoother distance distributions that better correlate with ground-truth similarity levels at the global scale. We attribute this behaviour to the fact that Spearman measures global rank agreement across all script pairs, whereas NDCG@10 focuses on the top of the ranking — and it is precisely this top-$k$ precision that our hybrid initialization improves.

\subsubsection{Secondary Metrics: Glyph-Level Retrieval.}

At the glyph level, Top-1 and Top-5 accuracy measure whether the nearest neighbour of a query glyph belongs to the correct character class in a 20-way 1-shot retrieval setting. While these metrics do not directly reflect the script-level objectives of our framework, they provide a complementary assessment of the discriminative quality of the learned representations. On Simple CNN and ResNet-50, our approach is competitive with or superior to all baselines. On Simple CNN, Ours (Teacher) achieves 88.0 on N20R1 and 98.75 on N20R5, surpassing both SupCon (85.5), BYOL (43.5), and Barlow Twins (77.75). On ResNet-50, Ours matches Barlow Twins on N20R5 (98.75 vs 100) while achieving a higher NDCG@10, confirming that our method does not sacrifice glyph-level discrimination for script-level coherence.

On ResNet-18 and ResNet-34, however, Barlow Twins and BYOL achieve higher Top-1 accuracy than our approach. We attribute this to the architecture mismatch between Stage 1 and Stage 2: mid-size ResNet backbones may not fully benefit from the SupCon initialization, as their higher capacity makes them more prone to drifting away from the teacher's structured representation space during the BYOL adaptation phase. Importantly, on these architectures, our method still achieves competitive NDCG@10 and recovers strong performance on Spearman, suggesting that the script-level geometry remains coherent even when glyph-level Top-1 is not maximised.

Finally, DINOv2-ViT-S/14, despite its 22M parameters, achieves modest results across all metrics in both frozen and fine-tuned configurations. This confirms that large vision transformers pre-trained on natural images do not transfer well to the highly specific domain of ancient script analysis, and underlines the importance of domain-adapted training strategies such as the one proposed in this work.

\begin{table*}[ht] 
\centering
\caption{Comparison of backbone architectures and training strategies on glyph- and script-level evaluation. We report 20-way 1-shot retrieval accuracy (N20R1/N20R5), script-level ranking quality (NDCG@10), and Spearman correlation with curated linguistic similarity levels. ``Teacher Type'' indicates whether an EMA teacher is used; here, the teacher is obtained by supervised contrastive pretraining followed by EMA self-distillation (SupCon$\rightarrow$EMA). Best results are in \textbf{bold}, second best are \underline{underlined}.}
\label{tab:backbone_comparison}
\resizebox{\textwidth}{!}{
\begin{tabular}{l|c|l|l|cccc}
\toprule
\textbf{Backbone} & \textbf{Params} & \textbf{Method} & \textbf{Teacher Type} & \textbf{N20R1} & \textbf{N20R5} & \textbf{NDCG10} & \textbf{Spearman} \\
\midrule
\multirow{5}{*}{\textbf{Simple CNN}} & \multirow{5}{*}{2.3M} 
& SupCon & - & 85.5 & \textbf{98.75} & 0.2781 & 0.548 \\
& & BYOL & EMA & 43.5 & 83.75 & 0.1932 & 0.545 \\
& & Barlow Twins & - & 77.75 & 97.75 & 0.2281 & \underline{0.639} \\
& & \textbf{Ours (Target)} & SupCon→EMA & \textbf{88} & \textbf{98.75} & \textbf{0.2930} & \textbf{0.640} \\
& & \textbf{Ours (Student)} & SupCon→EMA & \textbf{88} & \textbf{98.75} & \textbf{0.2930} & \underline{0.639} \\
\midrule
\multirow{5}{*}{\textbf{Siamese CNN}} & \multirow{5}{*}{8.7M} 
& SupCon & - & \underline{86.75} & \underline{99.5} & \textbf{0.3011} & \underline{0.631} \\
& & BYOL & EMA & 75.00 & 94.75  & \underline{0.2716} & \textbf{0.642} \\
& & Barlow Twins & - & \textbf{94.25} & \textbf{99.75} & 0.2665 & 0.536 \\
& & \textbf{Ours (Target)} & SupCon→EMA & 85 & 99 & 0.2677 & 0.594 \\
& & \textbf{Ours (Student)} & SupCon→EMA & 85 & 99 & 0.2677 & 0.594 \\\midrule
\multirow{5}{*}{\textbf{ResNet-18}} & \multirow{5}{*}{11.7M} 
& SupCon & - & 64 & 84.75 & 0.2480 & \underline{0.471} \\
& & BYOL & EMA & \underline{89.25} & \underline{99.50} & \underline{0.2882} & \textbf{0.617} \\
& & Barlow Twins & - & \textbf{93.75} & \textbf{99.75} & \textbf{0.3031} & 0.431 \\
& & \textbf{Ours (Target)} & SupCon→EMA & 63.25 & 86.75 & 0.2537 & 0.379 \\
& & \textbf{Ours (Student)} & SupCon→EMA & 63.5 & 85 & 0.2532 & 0.373 \\\midrule
\multirow{5}{*}{\textbf{ResNet-34}} & \multirow{5}{*}{21.8M} 
& SupCon & - & 73 & 94.5 & 0.2647 & 0.433 \\
& & BYOL & EMA & \underline{86.25} & \underline{98.00} & 0.2428 & \underline{0.445} \\
& & Barlow Twins & - & \textbf{93.00} & \textbf{99.5} & \underline{0.2895} & \textbf{0.482} \\
& & \textbf{Ours (Target)} & SupCon→EMA & 70.75 & 92.75 & 0.2518 & 0.383 \\
& & \textbf{Ours (Student)} & SupCon→EMA & 84.25 & 96.75 & \textbf{0.2927} & 0.388 \\\midrule
\multirow{5}{*}{\textbf{ResNet-50}} & \multirow{5}{*}{25.6M} 
& SupCon & - & 86.25 & 97.50 & 0.3078 & \underline{0.514} \\
& & BYOL & EMA & 87.50 & \underline{99.25} & 0.2708 & \textbf{0.622} \\
& & Barlow Twins & - & \textbf{93.75} & \textbf{100} & 0.2997 & 0.481 \\
& & \textbf{Ours (Target)} & SupCon→EMA & \underline{93.00} & 98.75 & \textbf{0.3178} & 0.424 \\
& & \textbf{Ours (Student)} & SupCon→EMA & \underline{93.00} & 98.75 & \textbf{0.3178} & 0.424 \\\midrule
\multirow{3}{*}{\textbf{DINOv2-ViT-S 14}} & \multirow{3}{*}{22M}
& Frozen & - & 43.75 & 74.75 & 0.2612 & 0.477 \\
& & Fine-tuned & - &  61 & 90.5 & 0.2366 & 0.609 \\
\bottomrule
\end{tabular}
}\label{table_results}
\end{table*}

\begin{figure}[h!]
  \centering
  \includegraphics[width=\columnwidth]{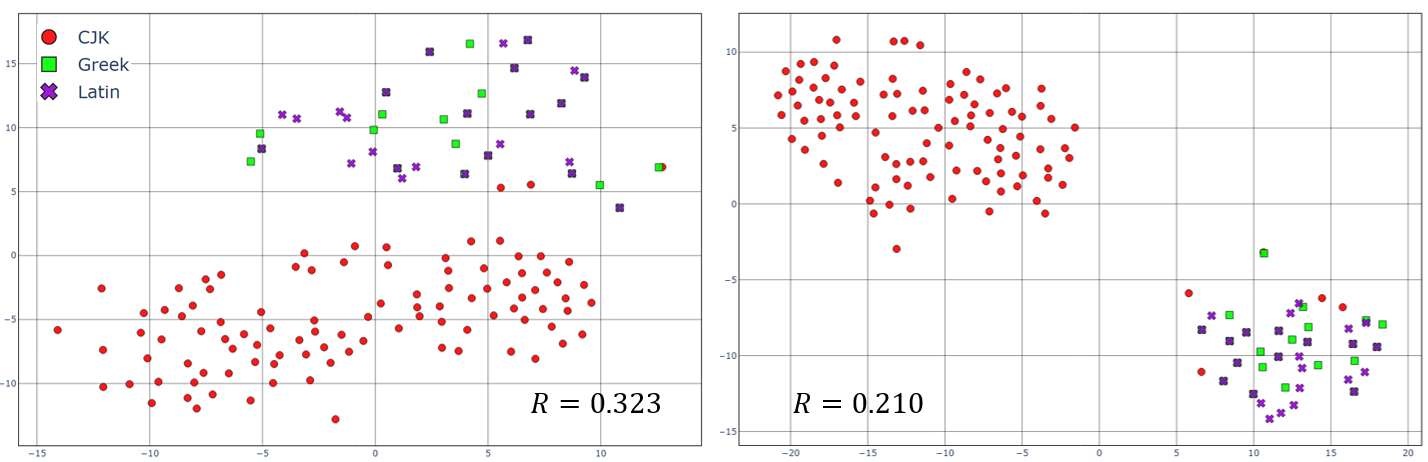}
    \caption{Two-dimensional t-SNE projections \cite{van2008visualizing} of glyph embeddings for the CJK, Greek and Latin scripts, shown on the left using the teacher model $f_{\phi}^*$ and on the right the student model $f_{\theta}$. Points correspond to glyph instances; colors/markers indicate script labels. The reported $R$ denotes the separability ratio \eqref{eq:sep_ratio}, illustrating how improved representation geometry yields more coherent cross-script structure.}
  \label{fig:t_sne}
\end{figure}

\subsubsection{Separability Ratio.}
To further assess the geometric quality of the learned embedding space, we introduce a separability ratio $\mathcal{R}$ that measures how much closer linguistically related scripts are compared to unrelated ones. Given three scripts, for instance Greek, Latin, and CJK (Chinese-Japanese-Korean) characters, where Greek and Latin share a common historical origin while CJK is linguistically independent, we define
\begin{equation} \label{eq:sep_ratio}
\mathcal{R} = \frac{d_s(\text{Greek}, \text{Latin})}{\frac{1}{2}\left[d_s(\text{CJK}, \text{Greek}) + d_s(\text{CJK}, \text{Latin})\right]}
\end{equation}
A lower value of $\mathcal{R}$ indicates that linguistically related scripts (Greek, Latin) are embedded proportionally closer to each other than to an unrelated script (CJK), reflecting a geometrically coherent organisation of the embedding space with respect to historical relationships. As shown in Figure~\ref{fig:t_sne}, the teacher $f_\phi^*$ yields $\mathcal{R} = 0.323$, while the student $f_\theta$ achieves $\mathcal{R} = 0.210$, a relative reduction of $35\%$. This result demonstrates that the unsupervised adaptation in Stage 2 does not merely compress the embedding space uniformly, but selectively accentuates historically grounded proximities, producing a geometry that better reflects the linguistic structure of writing systems.

\section{Conclusion}

We studied glyph and script similarity in a setting where supervision is fundamentally asymmetric: character identity can be safely supervised in invented alphabets, while cross-script relations in historical writing systems are often uncertain and should not be enforced through negative pairs. To address this, we proposed a two-stage framework that learns a discriminative prior with supervised contrastive training on labeled invented scripts, then adapts to historical scripts via teacher-initialized self-distillation without cross-script negatives.

Across five backbones and two datasets (Omniglot and our Unicode benchmark), this strategy consistently improves script-level ranking quality, achieving the best NDCG@10 overall while remaining competitive on 20-way 1-shot glyph retrieval. The separability ratio and t-SNE analyses further indicate that Stage~2 does not simply compress the teacher space, but sharpens historically grounded proximities, yielding a more coherent script geometry. Comparisons with BYOL, Barlow Twins, SupCon, and DINOv2 highlight that domain-adapted training is crucial for ancient script analysis, beyond generic off-the-shelf visual features.

Future work will leverage the resulting script distance as a building block for broader phylogenetic analyses of writing systems at worldwide scale, investigating whether the induced geometry can support tree- or network-based reconstructions of script lineages. In parallel, we plan to enrich training data with learned augmentations, using deep generative models (e.g., GAN-based synthesis) to emulate realistic handwritten variations and style drift beyond standard geometric transforms. Finally, beyond writing systems, the same two-stage principle, learn a discriminative prior where supervision is reliable, then adapt without imposing speculative negatives, could be transferred to other asymmetric settings where within-class identity is known but cross-category relations are uncertain, incomplete, or contested.

\bibliographystyle{splncs04}
\bibliography{main}

\end{document}